\documentclass{article}

\usepackage{graphicx,amsmath,amssymb} 
\usepackage{subfigure} 

\usepackage{natbib}

\usepackage{algorithm}
\usepackage{algorithmic}

\usepackage{hyperref}

\def \diag{\mbox{diag}}
\def \blkdiag{\mbox{blkdiag}}

\def \trace{\mbox{tr}}
\def \rank{\mbox{rank}}

\def \vec{\mbox{vec}}

\def\hbx{\hat{{\mathbf x}}}
\def\bx{{\mathbf x}}
\def\by{{\mathbf y}}
\def\bu{{\mathbf u}}
\def\bn{{\mathbf n}}

\def\bb{{\mathbf b}}

\newcommand{\cL}[0]{\mathcal{L}}
\newcommand{\cC}[0]{\mathcal{C}}
\newcommand{\cS}[0]{\mathcal{S}}

\newcommand{\rR}[0]{\mathbb{R}}
\def \blambda{\boldsymbol{\lambda}}
\def \bbeta{\boldsymbol{\beta}}

\usepackage[accepted]{icml2014} 

\usepackage[fontsize=9.8]{scrextend}

\icmltitlerunning{Convex Total Least Squares}

\begin{document} 

\twocolumn[
\icmltitle{Convex Total Least Squares}

\icmlauthor{Dmitry Malioutov}{dmalioutov@us.ibm.com}
\icmladdress{IBM Research,
            1101 Kitchawan Road, Yorktown Heights, NY 10598 USA}
\icmlauthor{Nikolai Slavov}{nslavov@alum.mit.edu}
\icmladdress{Departments of Physics and Biology, MIT,
            77 Massachusetts Avenue, Cambridge, MA 02139, USA}

\icmlkeywords{total least squares, convex relaxation, nuclear norm}

\vskip 0.3in
]

\begin{abstract} 
We study the total least squares (TLS) problem that generalizes least 
squares regression by allowing measurement errors in both dependent and 
independent variables. TLS is widely used in applied fields including computer 
vision, system identification and econometrics. The special case 
when all dependent and independent variables have the same level of uncorrelated 
Gaussian noise, known as ordinary TLS, can be solved by singular value 
decomposition (SVD). However, SVD cannot solve many important 
practical TLS problems with realistic noise structure, such as having varying 
measurement noise, known structure on the errors, or large outliers requiring 
robust error-norms. To solve such problems, we develop convex relaxation approaches 
for a general class of structured TLS (STLS). We show both theoretically 
and experimentally, that while the plain nuclear norm relaxation incurs
large approximation errors for STLS, the re-weighted nuclear norm approach is very 
effective, and achieves better accuracy on challenging STLS problems than popular 
non-convex solvers. We describe a fast solution based on augmented Lagrangian 
formulation, and apply our approach to an important class of biological problems 
that use population average measurements to infer cell-type and physiological-state 
specific expression levels that are very hard to measure directly.
\end{abstract} 

\section{Introduction}

Total least squares is a powerful generalization of ordinary least squares (LS)
which allows errors in the measured explanatory variables 
\cite{golub_van_loan_TLS}. It has become an indispensable tool in a variety of 
disciplines including chemometrics, system identification, astronomy, computer 
vision, and econometrics \cite{markovsky_local_STLS}. Consider a least squares 
problem $\by \approx X \bbeta$, where we would like to find coefficients 
$\bbeta$ to best predict the target vector $\by$ based on measured variables $X$. 
The usual assumption is that $X$ is known exactly, and that the errors come from 
i.i.d. additive Gaussian noise $\bn$: $\by = X \bbeta + \bn$. The LS problem has 
a simple closed-form solution by minimizing $\Vert \by - X \bbeta \Vert_2^2$ with 
respect to $\bbeta$.  In many applications not only $\by$ but also $X$ is
known only approximately, $X = X_0 + E_x$, where $X_0$ are the uncorrupted 
values, and $E_x$ are the unknown errors in observed variables. The total least 
squares (TLS) formulation, or errors in variables regression, tries to jointly minimize 
errors in $\by$ and in $X$ ($\ell_2$-norm of $\bn$ and Frobenius norm of $E_x$):
\begin{equation}
\label{eqn:plain_TLS}
\min_{\bn, E_x, \bbeta}  \Vert \bn \Vert_2^2 + \Vert E_x \Vert_F^2 ~~\mbox{where}~~ 
	\by = (X - E_x) \bbeta + \bn
\end{equation} 
While the optimization problem in this form is not convex, it can in fact be 
reformulated as finding the closest rank-deficient matrix to a given matrix, and 
solved in closed form via the singular value decomposition (SVD) \cite{golub_van_loan_TLS}.

Many error-in-variables problems of practical interest have additional information: 
for example, a subset of the entries in $X$ may be known exactly, we may know 
different entries with varying accuracy, and in general $X$ may exhibit a 
certain structure, e.g. block-diagonal, Toeplitz, or Hankel in system 
identification literature \cite{STLS_sys_id}. Furthermore, it is often important 
to use an error-norm robust to outliers, e.g. Huber loss or $\ell_1$-loss. 
Unfortunately, with rare exceptions\footnote{A closed form solution exists 
when subsets of columns are fully known; a Fourier transform based approach
can handle block-circulant errors $E_x$  \cite{ben_tal_block_circulant}.}, 
none of these problems allow an efficient solution, 
and the state of the art approach is to solve them by local optimization methods 
\cite{slra_Markovsky, Giannakis_sparse_STLS, srebro2003weighted}.
The only available guarantee
is typically the ability to reach a stationary point of the non-convex objective.

In this paper we propose a principled formulation for STLS 
based on convex relaxations of matrix rank. Our approach uses the re-weighted 
nuclear norm relaxation \cite{fazel_log_det} and is highly flexible: it can handle very general 
linear structure on errors, including arbitrary weights (changing noise for 
different entries), patterns of observed and unobserved errors, Toeplitz and 
Hankel structures, and even norms other than the 
Frobenius norm. The nuclear norm relaxation has been successfully used for a range
of machine learning problems involving rank constraints, including low-rank matrix 
completion, low-order system approximation, and robust 
PCA \cite{candes_iter_threshold, venkat_sparse_low_rank}. The STLS problem
is conceptually different in that we do not seek low-rank solutions, but 
on the contrary nearly full-rank solutions. We show both theoretically and 
experimentally that while the plain nuclear norm formulation incurs large approximation 
errors, these can be dramatically improved by using the re-weighted nuclear norm. 
We suggest fast first-order methods based on Augmented Lagrangian 
multipliers \cite{bertsekas_ALM_book} to compute the STLS solution. As part of ALM 
we derive new updates for the re-weighted nuclear-norm based on 
solving the Sylvester's equation, which can also be used for many other 
machine learning tasks relying on matrix-rank, including matrix completion 
and robust PCA. 

As a case study of our approach to STLS we consider an important application in 
biology, quantification of cellular heterogeneity \cite{slavov2011coupling}.  
We develop a new representation for the problem as a large structured linear 
system, and extend it to handle noise by a structured TLS problem with 
block-diagonal error structure. Experiments demonstrate the effectiveness of 
STLS in recovering physiological-state specific expression levels from aggregate 
measurements.

\subsection{Total Least Squares}
\label{S:TLS}

We first review the solution of ordinary TLS problems. We simplify the notation from 
(\ref{eqn:plain_TLS}): combining our noisy data 
$X$ and $\by$ into one matrix, $\bar{A} \triangleq [X ~-\by]$, and the errors into 
$E \triangleq [E_x ~-\bn]$ we have
\begin{equation}
	\min \Vert E \Vert_F^2  ~~\mbox{ where } (\bar{A} - E)
	\left[ \begin{matrix} \bbeta \\ 1 \end{matrix} \right] = 0.
\end{equation}

The matrix $\bar{A}$ is in general full-rank, and a solution can be obtained by 
finding a rank-deficient matrix closest to $\bar{A}$ in terms of the 
Frobenius norm. This finds smallest errors $E_x$ and $\bn$ such 
that $\by + \bn$ is in the range space of $X - E_x$. The closest rank-deficient 
matrix is simply obtained by computing the SVD, $\bar{A} = U S V^T$ and setting 
the smallest singular value to be zero. 

Structured TLS problems \cite{markovsky_local_STLS} allow more realistic errors 
$E_x$: with subsets of measurements that may be known exactly; weights reflecting 
different measurement noise for each entry; requiring linear structure 
of errors $E_x$ such as Toeplitz that is crucial in deconvolution 
problems in signal 
processing. Unfortunately, the SVD does not apply to any of these more general 
versions of TLS 
\cite{srebro2003weighted, markovsky_local_STLS}. Existing solutions to structured 
TLS problems formulate a non-convex optimization problem and attempt to solve it 
by local optimization \cite{slra_Markovsky} that suffers
from local optima and lack of guarantees on accuracy. We follow a different route and use
a convex relaxation for the STLS problem.

\section{STLS via a nuclear norm relaxation}
\label{S:STLS}

The STLS problem in a general form can be described as follows \cite{markovsky_local_STLS}. 
Using the notation in Section \ref{S:TLS}, suppose our observed matrix $\bar{A}$ is 
$M \times N$ with full column rank. We aim to find a nearby rank-deficient matrix $A$, 
$\rank(A) \le N-1$, where the errors $E$ have a certain linear structure:
\begin{eqnarray}
\label{eqn:STLS_hard}
\nonumber
\min \Vert W \odot E \Vert_F^2 ~\mbox{, where }~ \rank(A) \le N-1\\
A = \bar{A} - E ~\mbox{, and }~ \cL(E) = \bb
\end{eqnarray}
The key components here are the linear equalities that $E$ has to satisfy, 
$\cL(E) = \bb$. This notation represents a set of linear constraints 
$\trace(L_i^T E) = b_i$, for $i = 1, .., J$. In our application to cell 
heterogeneity quantification these constraints correspond to knowing certain 
entries of $A$ exactly, i.e. $E_{ij}=0$ for some 
subset of entries, while other entries vary freely. One may require other linear 
structure such as Toeplitz or Hankel. We also allow an element-wise 
weighting $W \odot E$, with $W_{i,j} \ge 0$ on 
the errors, as some observations may be measured with higher accuracy than others. 
Finally, while we focus on the Frobenius norm of the error, any other convex 
error metric, for example, mean absolute error, or robust Huber loss, could be 
used instead. The main difficulty in the formulation is posed by the non-convex 
rank constraint. The STLS problem is a special case of the structured low-rank 
approximation problem, where rank is exactly $N-1$ \cite{slra_Markovsky}. Next, 
we propose a tractable formulation for STLS based on convex relaxations of matrix 
rank.

We start by formulating the nuclear-norm relaxation for TLS and then improve upon 
it by using the re-weighted nuclear norm. The nuclear norm $\Vert A \Vert_*$ is 
a popular relaxation used to convexify rank constraints \cite{candes_iter_threshold}, 
and it is defined as the sum of the singular values of the matrix $A$, i.e. 
$\Vert A \Vert_* = \sum_i \sigma_i(A)$. It can be viewed as 
the $\ell_1$-norm of the singular value spectrum\footnote{For diagonal matrices
$A$ the nuclear norm is exactly equivalent to the $\ell_1$-norm of the diagonal 
elements.} favoring few non-zero singular values, i.e., matrices 
with low-rank. Our initial nuclear norm relaxation for the STLS problem is:
\begin{eqnarray}
\label{eqn:STLS_nuclear}
\nonumber
\min \Vert A \Vert_* + \alpha \Vert W \odot E \Vert_F^2 \mbox{ such that } \\
~~ A = \bar{A} - E ~\mbox{, and }~ \cL(E) = \bb
\end{eqnarray}
The parameter $\alpha$ balances error residuals vs. the nuclear norm (proxy for rank).
We chose the largest $\alpha$, i.e. smallest nuclear norm penalty, that still produces 
$\rank(A) \le N-1$. This can be achieved by a simple binary search over $\alpha$. 
In contrast to matrix completion and robust PCA, the STLS problem aims to find almost fully dense 
solutions with rank $N-1$, so it requires different analysis tools. We present 
theoretical analysis specifically for the STLS problem in Section \ref{S:rec_analysis}.  
Next, we describe the re-weighted nuclear norm, which, as we show in Section \ref{S:rec_analysis}, 
is better suited for the STLS problem than the plain nuclear norm.

\subsection{Reweighted nuclear norm and the log-determinant heuristic for rank}
\label{S:rw_NN}

A very effective improvement of the nuclear norm comes from re-weighting 
it \cite{fazel_log_det, karthik_mohan_reweightedNN} based on the log-determinant
heuristic for rank. To motivate it, we first describe a closely related approach in the 
vector case (where instead of searching for low-rank matrices one would like to find sparse 
vectors). Suppose that we seek a sparse solution to a general convex 
optimization problem. A popular approach penalizes the $\ell_1$-norm of the solution $\bx$
$\Vert \bx \Vert_1 = \sum_i |x_i|$ to encourage sparse solutions. A dramatic improvement 
in finding sparse signals can be obtained simply by using the weighted $\ell_1$-norm, 
i.e. $\sum_i w_i |x_i|$ with suitable positive weights $w_i$ \cite{candes_rev_l1} instead of a
plain $\ell_1$-norm. Ideally the weights would be based on the unknown signal, to provide 
a closer approximation to sparsity ($\ell_0$-norm)
by penalizing large elements less than small ones. A  practical solution first solves a 
problem involving the unweighted $\ell_1$-norm, and uses the solution $\hbx$ to define 
the weights $w_i = \frac{1}{\delta + |\hat{x}_i|}$, with $\delta$ a small positive constant. 
This iterative approach can be seen as an iterative local linearization of the concave 
$\log$-penalty for sparsity, $\sum_i \log (\delta + |x_i|)$ 
\cite{fazel_log_det, candes_rev_l1}. In both empirical and emerging theoretical 
studies\cite{needell2009noisy, hassibi2009weighted} re-weighting the $\ell_1$-norm has 
been shown to provide a tighter relaxation of sparsity.

In a similar way, the re-weighted nuclear norm tries to penalize large singular 
values less than small ones by introducing positive weights. There is an analogous 
direct connection to the iterative linearization for the concave log-det relaxation 
of rank \cite{karthik_mohan_reweightedNN}. Recall that the problem of minimizing the 
nuclear norm subject to convex set constraints $\cC$,
\begin{equation}
\min \Vert A \Vert_* ~ \mbox{ such that } ~ A \in \cC , 
\end{equation}
has a semi-definite programming (SDP) representation
\cite{fazel_log_det}. Introducing auxiliary symmetric 
p.s.d. matrix variables $Y, Z \succeq 0$, we rewrite it as:
\begin{equation}
\min_{A, Y, Z} \trace(Y) + \trace(Z) ~~~\mbox{ s.t.} 
\left[ \begin{matrix} Y & A \\ A^T & Z \end{matrix} \right] \succeq 0, ~~ A \in \cC
\end{equation}
Instead of using the convex nuclear norm relaxation, it has been suggested
to use the concave log-det approximation to rank:
\begin{eqnarray}
\nonumber
\min_{A, Y, Z} \log \det (Y + \delta I ) + \log \det(Z + \delta I) \\
~~~\mbox{ s.t.} 
\left[ \begin{matrix} Y & A \\ A^T & Z \end{matrix} \right] \succeq 0, ~~ A \in \cC
\end{eqnarray}
Here $I$ is the identity matrix and $\delta$ is a small positive constant. 
The log-det relaxation provides a closer approximation to rank than the nuclear norm, but 
it is more challenging to optimize. By iteratively linearizing this objective one 
obtains a sequence of weighted nuclear-norm problems \cite{karthik_mohan_reweightedNN}:
\begin{eqnarray}
\nonumber
\min_{A, Y, Z} \trace ((Y^k + \delta I )^{-1} Y) + \trace ((Z^k + \delta I)^{-1} Z) \\
~~~\mbox{ s.t.} 
\left[ \begin{matrix} Y & A \\ A^T & Z \end{matrix} \right] \succeq 0, ~~ A \in \cC
\end{eqnarray}
where $Y^k, Z^k$ are obtained from the previous iteration, 
and $Y^0, Z^0$ are initialized as $I$. Let $W_1^k = (Y^k + \delta I)^{-1/2}$ and 
$W_2^k = (Z^k + \delta I)^{-1/2}$ then the problem is equivalent to a weighted 
nuclear norm optimization in each iteration $k$:
\begin{eqnarray}
\label{eqn:rw_trace_norm}
\min_{A, Y, Z} \Vert W_1^k A W_2^k \Vert_* ~~~\mbox{ s.t.} ~~A \in \cC
\end{eqnarray} 
The re-weighted nuclear norm approach iteratively solves convex weighted nuclear norm 
problems in (\ref{eqn:rw_trace_norm}):

\paragraph{ Re-weighted nuclear norm algorithm:\\} Initialize: $k=0$, $W_1^0 = W_2^0 = I$.
\begin{itemize}
\item[(1)] Solve the weighted NN problem in (\ref{eqn:rw_trace_norm}) to get $A^{k+1}$.
\item[(2)] Compute the SVD: $W_1^k A^{k+1} W_2^k = U \Sigma V^T$, and set
$Y^{k+1} = (W_1^k)^{-1} U \Sigma U^T  (W_1^k)^{-1}~~$ and \\$~~Z^{k+1} = (W_2^k)^{-1} V \Sigma V^T  (W_2^k)^{-1}$.
\item[(3)] Set $W_1^k = (Y^k + \delta I)^{-1/2}$ and $W_2^k = (Z^k + \delta I)^{-1/2}$.
\end{itemize}

There are various ways to solve the plain and weighted nuclear norm STLS 
formulations, including interior-point methods \cite{TohTT1999} and iterative 
thresholding \cite{candes_iter_threshold}. In the next section we focus on augmented 
Lagrangian methods (ALM) \cite{bertsekas_ALM_book} which allow fast convergence 
without using computationally expensive second-order information.

\section{Fast computation via ALM}
\label{S:ALM}

While the weighted nuclear norm problem in (\ref{eqn:rw_trace_norm})
can be solved via an interior point method, it is computationally 
expensive even for modest size data because of the need 
to compute Hessians. We develop an effective first-order approach for 
STLS based on the augmented Lagrangian multiplier (ALM) method 
\cite{bertsekas_ALM_book,yi_ma_ALM}. Consider a general equality 
constrained optimization problem:
\begin{equation}
\label{eqn:eq_constr_opt}
\min_x f(\bx) ~~\mbox{such that}~~~  h(\bx) = 0.
\end{equation}
ALM first defines an augmented Lagrangian function:
\begin{equation}
\label{eqn:aug_lag}
L(\bx, \blambda, \mu) = f(\bx)~ + ~\blambda^T h(\bx)~ + ~\frac{\mu}{2} \Vert h(\bx) \Vert_2^2
\end{equation}
The augmented Lagrangian method alternates optimization over $\bx$ 
with updates of $\blambda$ for an increasing sequence of $\mu_k$. The 
motivation is that either if $\blambda$ is near the optimal dual solution 
for (\ref{eqn:eq_constr_opt}), or, if $\mu$ is large enough, then the 
solution to (\ref{eqn:aug_lag}) approaches the global minimum of 
(\ref{eqn:eq_constr_opt}). When $f$ and  $h$ are both continuously 
differentiable, if $\mu_k$ is an increasing sequence, the solution converges 
$Q$-linearly to the optimal one \cite{bertsekas_ALM_book}. The work of 
\cite{yi_ma_ALM} extended the analysis to allow objective functions involving 
nuclear-norm terms. The ALM method iterates the following steps:

\paragraph{Augmented Lagrangian Multiplier method}
\begin{itemize}
\item[(1)] $\bx_{k+1} = \arg \min_{\bx} L(\bx, \blambda_k, \mu_k)$
\item[(2)] $\blambda_{k+1} = \blambda_k + \mu_k h(\bx_{k+1})$
\item[(3)] Update $\mu_k \to \mu_{k+1}$ (we use $\mu_k = a^k$ with $a > 1$).
\end{itemize}
Next, we derive an ALM algorithm for nuclear-norm STLS and extend it to use
reweighted nuclear norms based on a solution of the Sylvester's equations.

\subsection{ALM for nuclear-norm STLS}
\label{S:ALM_for_STLS}

We would like to solve the problem:
\begin{eqnarray}
\label{eqn:STLS_nuclear_v2}
\min  \Vert A \Vert_* +  \alpha \Vert E \Vert_F^2,  ~~\mbox { such that }~~\\ 
\nonumber \bar{A} = A + E, ~~\mbox{ and }~~ \cL(E) = \bb
\end{eqnarray}

To view it as (\ref{eqn:eq_constr_opt}) we have $f(x) = \Vert A \Vert_* +  \alpha \Vert E \Vert_F^2$ and 
$h(x) = \{ \bar{A} - A - E, \cL(E) - \bb \}$. Using $\Lambda$ as our matrix Lagrangian 
multiplier, the augmented Lagrangian is:
{\footnotesize
\begin{equation}
\label{eqn:aug_lag_NN}
\min_{E: \cL(E) = \bb} \Vert A \Vert_* +  \alpha \Vert E \Vert_F^2 + 
\trace(\Lambda^T (\bar{A} - A - E )) + \frac{\mu}{2} \Vert \bar{A} - A - E \Vert_F^2.
\end{equation} 
}
Instead of a full optimization over $\bx = (E, A)$, we use coordinate 
descent which alternates optimizing over each matrix variable holding the 
other fixed. We do not wait for the coordinate descent to converge at each 
ALM step, but rather update $\Lambda$ 
and $\mu$ after a single iteration, following the inexact ALM algorithm in 
\cite{yi_ma_ALM}\footnote{This is closely related to the popular 
alternating direction of multipliers methods \cite{boyd2011distributed}.}. Finally, instead 
of relaxing the constraint $\cL(E) = \bb$, we keep the constrained form, and follow
each step by a projection \cite{bertsekas_ALM_book}. 

The minimum of (\ref{eqn:aug_lag_NN}) over $A$ is obtained by the singular 
value thresholding operation \cite{candes_iter_threshold}:
\begin{equation}
	\label{eqn:soft_thresh}
	A_{k+1} = \cS_{\mu^{-1}} \left( \bar{A} - E_k + \mu_k^{-1} \Lambda_k   \right)
\end{equation}
where $\cS_{\gamma}(Z)$ soft-thresholds the singular values of $Z = U S V^T$, i.e. 
$\tilde{S} = \max(S - \gamma, 0)$ to obtain $\hat{Z} = U \tilde{S} V^T$.

The minimum of (\ref{eqn:aug_lag_NN}) over $E$ is obtained by setting the gradient  
with respect to $E$ to zero, followed by a projection\footnote{For many constraints of interest
this projection is highly efficient: when the constraint fixes some entries $E_{ij} = 0$, 
projection simply re-sets these entries to zero. Projection onto Toeplitz structure simply 
takes an average along each diagonal, e.t.c} onto the affine space defined by $\cL(E) = \bb$:
\begin{eqnarray}
\label{eqn:E_update_NN}
\nonumber
	\tilde{E}_{k+1} = \frac{1}{2 \alpha + \mu_k} \left( \Lambda_k + \mu_k ( \bar{A} - A) \right) \\
	~~~~\mbox{ and }~~~~
	E_{k+1} = \Pi_{E: \cL(E) = \bb} \tilde{E}_{k+1}
\end{eqnarray}



\subsection{ALM for re-weighted nuclear-norm STLS}

To use the log-determinant heuristic, i.e., the re-weighted nuclear norm approach,
we need to solve the weighted nuclear norm subproblems:
\begin{eqnarray}
\label{eqn:STLS_nuclear_v3}
\min  \Vert W_1 A W_2 \Vert_* +  \alpha \Vert E \Vert_F^2 ~~\mbox{ where } \\ \nonumber
\bar{A} = A + E ~\mbox{, and }~ \cL(E) = \bb
\end{eqnarray}

There is no known analytic thresholding solution for the weighted nuclear norm, 
so instead we follow \cite{yi_ma_rew_NN} to create a new variable $D = W_1 A W_2$ and 
add this definition as an additional linear constraint:
\begin{eqnarray}
\label{eqn:STLS_nuclear_v4}
\min  \Vert D \Vert_* +  \alpha \Vert E \Vert_F^2 ~~\mbox{ where } \\ \nonumber
\bar{A} = A + E , ~D =  W_1 A W_2  ~\mbox{, and }~ \cL(E) = \bb
\end{eqnarray}

Now we have two Lagrangian multipliers $\Lambda_1$ and $\Lambda_2$ 
and the augmented Lagrangian is 
\begin{eqnarray}
\nonumber
\min_{E : \cL(E) = \bb} \Vert D \Vert_* +  \alpha \Vert E \Vert_F^2 + 
\trace(\Lambda_1^T ~(\bar{A} - A - E)) + \\
 \trace (\Lambda_2^T (~D -  W_1 A W_2)) +  
\nonumber \\ \frac{\mu}{2} \Vert \bar{A} - A - E \Vert_F^2 + 
\frac{\mu}{2} \Vert D -  W_1 A W_2 \Vert_F^2
\end{eqnarray}
We again follow an ALM strategy, optimizing over $D, E, A$ separately followed by updates
of $\Lambda_1, \Lambda_2$ and $\mu$. Note that \cite{ALM_rew_NN_chinese_guys} considered a 
strategy for minimizing re-weighted nuclear norms for matrix completion, but instead of 
using exact minimization over $A$, they took a step in the gradient direction. We derive
the exact update, which turns out to be very efficient via a Sylvester equation formulation. 
The updates over $D$ and over $E$ look similar to the un-reweighted case. Taking a derivative 
with respect to $A$ we obtain 
a linear system of equations in an unusual form:
$-\Lambda_1 - W_y \Lambda_2 W_Z - \mu (\bar{A} - A - E) - \mu W_1 ( D - W_1 A W_2) W_2 = 0$.
Rewriting it, we obtain:
\begin{equation}
\label{eqn:sylvester}
A + W_1^2 A W_2^2 = \frac{1}{\mu_k} \left( \Lambda_1 + W_1 \Lambda_2 W_2 \right) + 
	(\bar{A} - E) + W_1 D W_2
\end{equation}
we can see that it is in the form of Sylvester equation arising in discrete Lyapunov systems
\cite{kailath1980linear}:
\begin{equation}
A + B_1 A B_2 = C
\end{equation}
where $A$ is the unknown, and $B_1, B_2, C$ are coefficient matrices. An efficient solution
is described in \cite{bartels_stewart}. These ALM steps for reweighted nuclear norm STLS 
are summarized in Algorithm \ref{alg:RW_NN_STLS}.

\begin{algorithm}[tb]
   \caption{ALM for weighted NN-STLS}
   \label{alg:RW_NN_STLS}
\begin{algorithmic}
   \STATE {\bfseries Input:} $\bar{A}$, $W_1$, $W_2$, $\alpha$
   \REPEAT
   \STATE $\bullet$ Update $D$ via soft-thresholding:\\
	$~~D_{k+1} = \cS_{\mu_k^{-1}} \left( W_1 A W_2 - 1/\mu_k \Lambda_2^k   \right)$.
   \STATE $\bullet$ Update $E$ as in (\ref{eqn:E_update_NN}).
   \STATE $\bullet$ Solve Sylvester system for $A$ in (\ref{eqn:sylvester}).
   \STATE $\bullet$ Update $\Lambda_1^{k+1} = \Lambda_1^k + \mu_k (\bar{A} - A - E)$, \\
$~~\Lambda_2^{k+1} = \Lambda_2^k + \mu_k (D - W_1 A W_2)$ and $\mu_k \to \mu_{k+1}$.
   \UNTIL{$convergence$}
\end{algorithmic}
\end{algorithm}
\vspace{-0.05in}

To obtain the full algorithm for STLS, we combine the above algorithm with steps 
of re-weighting the nuclear norm and the binary search over $\alpha$ as described in 
Section \ref{S:rw_NN}. We use it for experiments in Section \ref{S:exp}. A faster
algorithm that avoids the need for a binary search will be presented in a future publication. 

\section{Accuracy analysis for STLS}
\label{S:rec_analysis}

In context of matrix completion and robust PCA, the nuclear norm relaxation 
has strong theoretical accuracy guarantees \cite{lowrank_guarantee_Recht, 
venkat_sparse_low_rank}. We now study accuracy guarantees for the STLS 
problem via the nuclear norm and the reweighted nuclear norm approaches. 
The analysis is conducted in the plain TLS setting, where the optimal solution 
is available via the SVD, and it gives valuable insight into the accuracy of 
our approach for the much harder STLS problem. In particular, we quantify 
the dramatic benefit of using reweighting. In this section we study
a simplification of our STLS algorithm, where we set the 
regularization parameter $\alpha$ once and do not update it through the 
iterations. The full adaptive approach from Section \ref{S:rw_NN} is 
analyzed in the addendum to this paper where we show that it 
can in fact recover the exact SVD solution for plain TLS.

\begin{figure*}[t]
\begin{center}
 \begin{tabular}{ cc }
      \parbox[l]{2.5in}{
      \centerline{\includegraphics[width=2.5in]{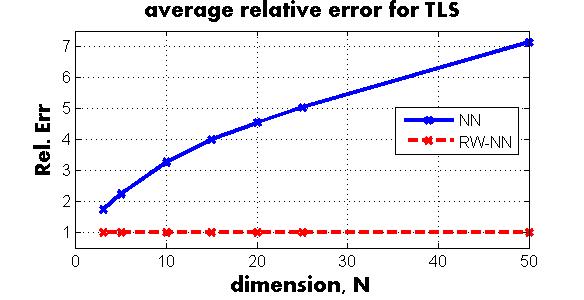}}
	} &
       \parbox[l]{2.5in}{
      \centerline{\includegraphics[width=2.5in]{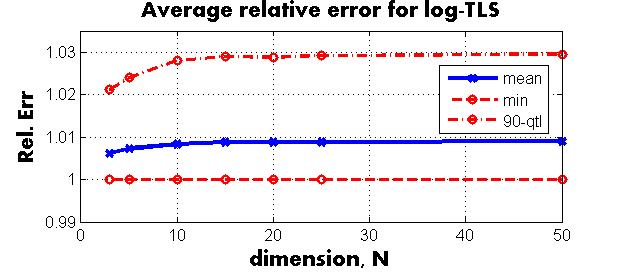}}
	}
    \end{tabular}
    \caption{\label{fig:RW_NN_TLS} RW-NN for plain TLS: (a) relative error 
	in Frobenius norm for NN and full RW-NN (avg. over 100 trials) w.r.t. SVD. (b) 
	Min, average, and $90 \%$-quantile relative error for the simplified non-adaptive log-det STLS 
	relative to SVD.}
\end{center}
\vspace{-0.2in}
\end{figure*}

We first consider the problem $\min \Vert A - \bar{A} \Vert^2_F$ such that 
$\rank(A) \le N-1$. For the exact solution via the SVD, the minimum 
approximation error is simply the square of the last singular value 
$Err_{SVD} = \Vert \hat{A}_{SVD} - \bar{A} \Vert^2_F = \sigma_N^2$. The nuclear-norm 
approximation will have a higher error. We solve 
$\min \Vert A - \bar{A} \Vert^2_F + \alpha \Vert A \Vert_*$ 
for the smallest choice of $\alpha$ that makes $A$ rank-deficient. A 
closed form solution for $A$ is the soft-thresholding operation with 
$\alpha = \sigma_N$. It subtracts $\alpha$ from all the singular values, 
making the error $Err_{\mbox{nn}} = N \sigma_N^2$. While it is bounded, 
this is a substantial increase from the SVD solution.
Using the log-det heuristic, we obtain much tighter accuracy guarantees
even when we fix $\alpha$, and do not update it during re-weighting. Let 
$a_i = \frac{\sigma_i} {\sigma_N}$, the ratio of the $i$-th and the 
smallest singular values. In the appendix using `log-thresholding' we derive that  
\begin{eqnarray}
	Err_{\mbox{rw-nn}} \approx \sigma_N^2 \left( 1 + \frac{1}{2} \sum_{i < N} ( a_i - \sqrt{a_i^2 - 1} )^2)\right) \\ \nonumber  \le \sigma_N^2 (1 + \frac{1}{2} \sum_{i < N} \frac{1}{a_i^2}). 
\end{eqnarray}
For larger singular values the approximation is much more accurate than for 
the smallest ones. In contrast, for the plain nuclear norm approach, the
errors are equally bad for the largest and smallest singular values. 
Considering that natural signals (and singular value 
spectra) often exhibit fast decay (exponential or power-law decay), we can quantify the 
improvement. Suppose that the singular values have exponential 
decay, $\sigma_i = \sigma_N a^{N-i}$, with $a > 1$, or power-law decay 
$\sigma_i = \sigma_N (N-i+1)^p$. The approximation errors are 
$Err_{exp} = \sigma_N^2 \left( 1 + \frac{1}{2} \sum_{i < N} (a^i - \sqrt{a^{2 i} - 1})^2 \right)$ and 
$Err_p = \sigma_N^2 \left( 1 + \frac{1}{2} \sum_{i < N} (i^p - \sqrt{i^{2 p} - 1})^2 \right)$
respectively. For exponential decay, if $N=100$, and $a = 1.1$, the 
approximation error is $1.84 ~\sigma_N^2$ for our approach, and 
$N \sigma_N^2 = 100 ~\sigma_N^2$, for the nuclear norm relaxation. 
This is a dramatic improvement in approximation, that strongly supports 
using the log-det heuristic over the nuclear norm for approximating matrix 
rank!

\section{Experimental Results}
\label{S:exp}

Our first experiment considers plain TLS, where we know the optimal solution 
via the SVD. We evaluate the accuracy of the nuclear norm (NN) and two flavors
of the reweighted nuclear norm algorithm: the full adaptive one 
described in Section \ref{S:rw_NN}, which we will refer to as (RW-NN), and 
the simplified approach with fixed $\alpha$ as described in Section 
\ref{S:rec_analysis} (log-det).

We simulate random i.i.d. Gaussian matrices $A$ of size $N \times N$, use a 
maximum of $3$ re-weightings in RW-NN, and update the ALM parameter $\mu_k$ 
as $\mu_k = 1.05^k$. We plot the relative error of NN-TLS with respect to exact 
TLS via SVD, i.e. the norm of error (w.o. squaring) for NN-TLS divided by the norm 
of error for TLS. We compare it to the relative error for full RW-NN TLS (again with 
respect to exact TLS) in Figure \ref{fig:RW_NN_TLS} (a). The results are averaged 
over $100$ independent trials. 

The NN solution, as we expect, is a factor of $\sqrt{N}$ worse than TLS in 
Frobenius norm. The full RW-NN always recovers 
the optimal TLS solution, i.e. the relative error is exactly $1$, as we establish 
in the addendum. The simplified non-adaptive log-det STLS in Figure \ref{fig:RW_NN_TLS} 
(b) is almost as good as the adaptive: the average error is only about $1 \%$ 
higher than exact TLS, dramatically better than $\sqrt{N}$ for plain NN. These 
empirical results agree with our theoretical analysis in Section \ref{S:rec_analysis}.

\begin{figure*}[t]
\begin{center}
 \begin{tabular}{ cc }
      \parbox[l]{2.5in}{
        \centerline{\includegraphics[width=2.5in]{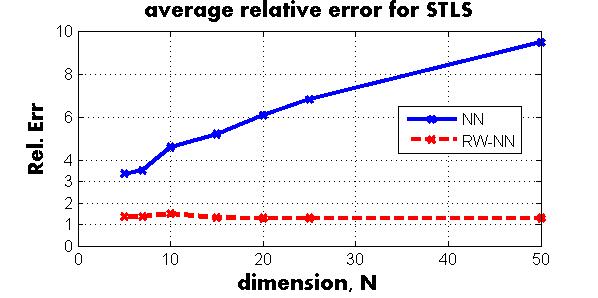}}
		} &
      \parbox[l]{2.5in}{
        \centerline{\includegraphics[width=2.5in]{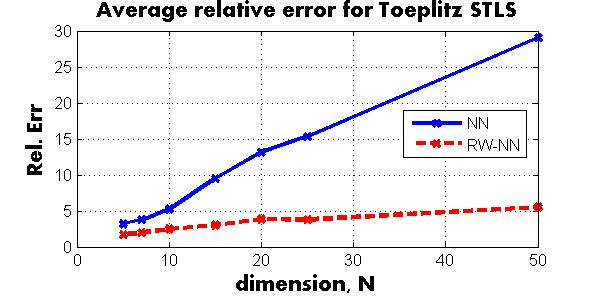}}
		}
    \end{tabular}
    \caption{\label{fig:RW_NN_STLS} Relative error in Frobenius norm for NN and RW-NN (avg. over 
	100 trials). The SVD solution is infeasible, providing a lower bound $1$ 
	on the error. (a) STLS with some observed entries (b) STLS with Toeplitz structure. }
\end{center}
\vspace{-0.2in}
\end{figure*}

Next, we compare NN and RW-NN for a structured TLS problem with a 
pattern of entries in $E$ fixed at $0$ (entries are fixed independently
with probability $0.5$). This is a practically important case where the 
entries of $E$ fixed at zero represent exact measurements while allowing other
entries to have noisy measurements. The solution of plain TLS via SVD ignores the
constraints and is infeasible for this structured problem. We still compute the 
relative error with respect to exact TLS to quantify the increase in error 
needed to obey the imposed structure. Again, in Figure \ref{fig:RW_NN_STLS} 
(a) we can see that the RW-NN solution provides much better accuracy than NN, and not 
far worse than $1$, the infeasible lower-bound given by plain TLS.

Next we consider Toeplitz structured errors, which means that the matrix
$E$ is constant on the diagonals:
\begin{equation}
\left[ \begin{matrix} 
e_{1} & e_{2} & e_{3} & ...\\
e_{0} & e_{1} & e_{2} & ...\\
e_{-1} & e_{0} & e_{1} & ...\\
... & ... & ... & ...
\end{matrix} \right]
\end{equation}
Toeplitz structure arises in time-series modeling, analysis of linear systems,
and system identification, as the convolution operation can be represented as a 
multiplication by a Toeplitz matrix \cite{kailath1980linear}. We simulate the 
Toeplitz entries at the start of each diagonal as i.i.d. Gaussian. The Toeplitz 
structure is quite restrictive, with only $M+N-1$ degrees of freedom  instead 
of $O(M N)$. However, we reach a similar conclusion as we had before: RW-NN 
solution provides much better accuracy than NN, much closer to the infeasible 
lower-bound via plain TLS. We show the results for STLS with Toeplitz 
structure in Figure \ref{fig:RW_NN_STLS}(b). 

\begin{figure}[t]
      \centerline{\includegraphics[width=2.25in]{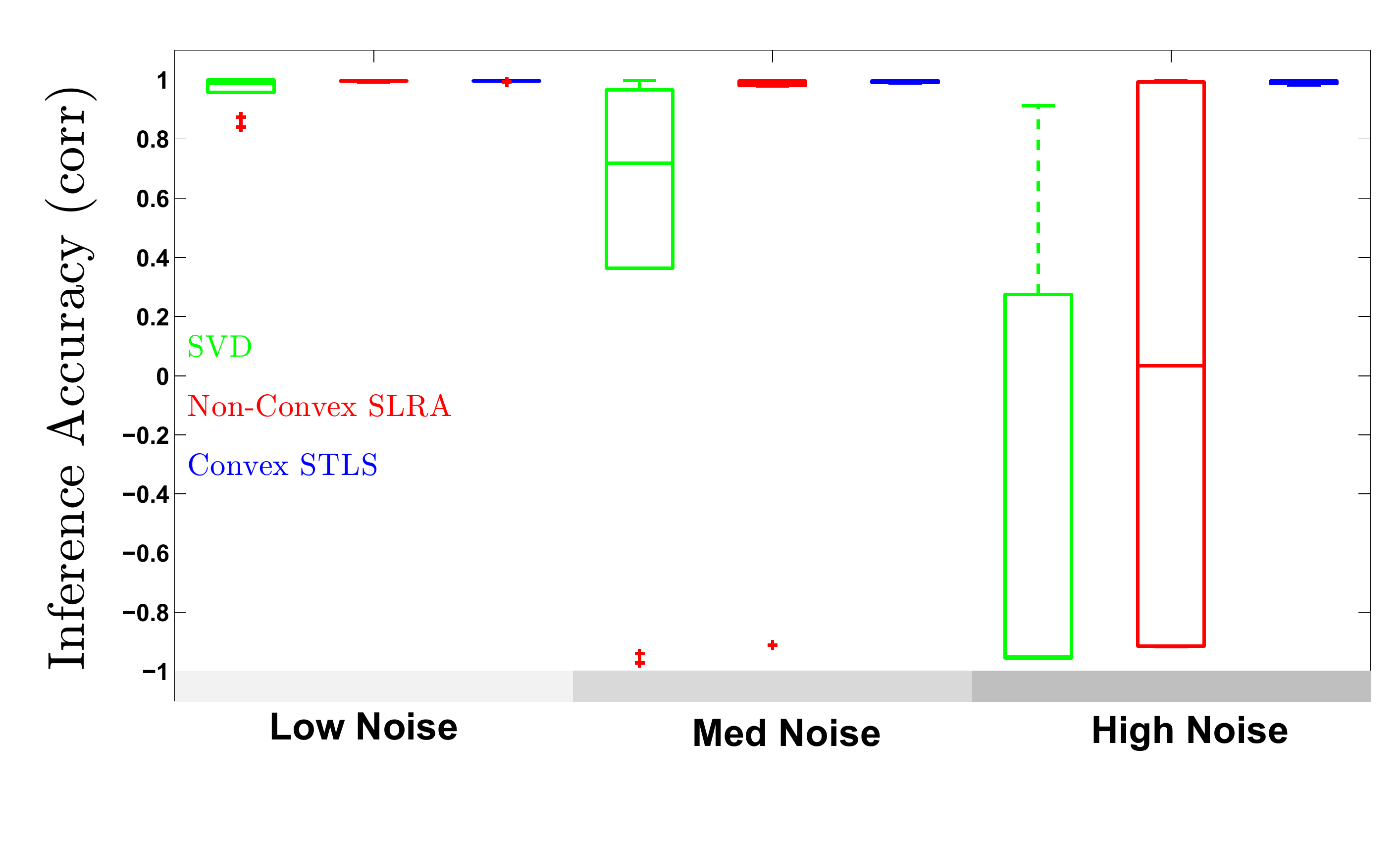}} 
    \caption{\label{fig:slra_example} Comparison of convex STLS (RW-NN) with 
	non-convex SLRA and SVD on problems with large outliers. At low-noise 
	all the solvers are accurate, but only RW-NN remains accurate at high noise. 
	We measure accuracy by correlation with the exact solution. 
	We present the distribution of correlations over $100$ trials as a boxplot.}
	\vspace{-0.2in}
\end{figure}

Finally, we compare our re-weighted nuclear norm approach to the 
latest widely used non-convex solver for STLS, SLRA \cite{slra_Markovsky}. The 
success of non-convex solvers depends heavily on a good initialization. We 
consider a problem with a block-diagonal structure where some entries are 
corrupted by large outliers. The weights on these entries are set to be very small, 
so an ideal STLS solver should find the solution while minimizing the influence of the 
outliers. Figure \ref{fig:slra_example} shows that for moderate levels of outliers, 
both RW-NN and the non-convex SLRA approach find very accurate solutions. However, for 
larger levels of noise, while RW-NN continues to have good performance, 
the accuracy of SLRA plummets, presumably due to the difficulty of finding a 
good initialization. For this setting we know the exact solution without outliers, 
and we measure accuracy by correlation (i.e. cosine of subspace angles) of the 
recovered and the exact STLS nullspaces, averaged over 100 trials.

\subsection{Quantification of cellular heterogeneity}
\label{S:real_world_problem}
We now demonstrate the utility of STLS for solving a broad and important class of 
problems arising in biology, namely inferring heterogeneity in biological systems. 
Most biological systems (such as human cancers, tissues, the human microbiome and 
other microbial communities) are mixtures of cells in different physiological 
states or even different cell types. While the primary biomedical interest is in 
characterizing the different cell types and physiological states, experimental 
approaches can typically measure only the population average across physiological states. 
Our aim is to combine these readily available population-average measurements and use 
STLS to infer the distribution of cells across distinct physiological states.

We consider a cell culture containing cells in $K$ distinct physiological states,
such as phases of cell growth or division cycles \cite{slavov2011metabolic,Slavov_emc}.
As the growth rate of the culture changes, the fraction of cells in each physiological 
state changes. This fractional change is of primary interest but it is often too 
expensive or even technologically impossible to measure directly. Since the 
cells cannot be easily separated we consider the 
general case when we know $M$ indicator genes (such as cyclins) that are either 
present or absent in $K$ distinct physiological states, $S \in \rR^{M \times K}$. 
Existing methods for high-throughput measurements of mRNA levels, such as DNA microarrays 
and RNA-seq, can quantify relative changes of mRNA levels across different conditions but cannot 
accurately quantify the ratios between different mRNAs, i.e., depending 
on chemical composition and the physical properties, each RNA has its own normalization scaler 
accounting for biases such as GC (guanine-cytosine) content. To avoid such biases we explicitly 
scale the measured relative expression levels $X \in \mathbb R^{M \times N}$ 
by an unknown positive diagonal matrix $Z = diag(z)$. The goal is to find $U \in \rR^{K \times N}$, 
the fraction of cells across the $K$ physiological states  for each of $N$ different conditions, 
such as different steady-state growth rates. Mathematically the problem is:
\begin{equation}
\label{eqn:problem_v1}
X = Z S U, 
\end{equation}
where we aim to recover the decomposition {\em up-to scaling} 
knowing $X$ and $S$ only. We now study conditions for identifiability without 
noise, and extend it to a structured TLS problem in presence of noise.


\paragraph{Linear Algebraic solution}
\label{S:lin_alg_sol}

We define $\blambda = [\frac{1}{z_1}, ... , \frac{1}{z_M}]$, and 
$\Lambda = \diag( \blambda)$. Thus $\Lambda = Z^{-1}$. We now have to 
find $\Lambda$ and $U$:
\begin{equation}
\label{eqn:problem_v2}
\Lambda X = S U, 
\end{equation}
and both unknowns enter the equations linearly. We transpose both sides 
and move everything to one side to get: $U^T S^T - X^T \Lambda = 0$.
Now let us stack columns of $U^T$, i.e. rows of $U$ into a vector, 
$\bu = \vec(U^T)$. Then $\vec(U^T S^T) = (S \otimes I) \bu$, where
$\otimes$ stands for the Kronecker product. Similarly defining
a block-diagonal matrix $\blkdiag(X^T)$, with columns of $X^T$ 
(i.e. rows of $X$) in diagonal blocks. This way $X^T \Lambda = \blkdiag(X^T) 
\blambda$. Combining this together we have:
\begin{eqnarray}
\label{eqn:problem_v4}
\left[ (S \otimes I),~  -\blkdiag(X^T)  \right] \left[\begin{matrix}\bu \\ 
\blambda\end{matrix} \right] = 0
\end{eqnarray}
Any vector in the nullspace of $A \triangleq \left[ (S \otimes I)~  -\blkdiag(X^T)  \right]$ is 
a solution to this problem. If we have a single vector in the nullspace
of $A$, then we have a unique solution up-to scaling.

\paragraph{Noisy case: structured Total Least Squares approach}
\label{S:noisy_STLS}

When the observation matrix $X$ is corrupted by noise, it is no longer 
low-rank. The structured matrix $A$ in (\ref{eqn:problem_v4}) will only 
have a trivial null-space. Furthermore, the simple approach of setting  
the smallest singular value to zero will not work because it ignores 
the structure of the compound matrix $A$. The errors in $X$ correspond 
to errors in the block-diagonal portions 
of the right part of $A$. Other entries are known exactly.
This is precisely the realm of structured total least squares (STLS) 
that we explored in Section \ref{S:STLS}. 

We will now experimentally apply our reweighted nuclear norm approach for 
STLS for the cell heterogeneity quantification problem to demonstrate the 
inference of the fractions of cells  in different physiological states. 
We use experimentally measured levels of $14$ genes, five expressed in HOC 
phase, six expressed in LOC phase, and three in both phases, across $6$ 
exponentially growing yeast cultures at different growth rates. The resulting
$A$ matrix in (\ref{eqn:problem_v4}) is $84 \times 26$. Our algorithm infers 
the fraction of cells in HOC and in LOC phase, up to a scalar factor, in 
close agreement with expectations from physical measurements in synchronized 
cultures \cite{slavov2011metabolic, slavov2011coupling}. Thus we can extend 
the observed trend to asynchronous cultures where this fraction is very hard to 
measure experimentally. Such analysis can empower research on cancer 
heterogeneity that is a major obstacle to effective cancer therapies. This 
modest size experiment provides a proof of concept and we are pursuing 
applications to more complex biological systems.

\section{Appendix: Error analysis for re-weighted STLS}

To gain insight into the re-weighted nuclear norm we consider the diagonal 
case first, where $A = \mbox{diag} (\bx)$. The diagonal matrix case penalizing rank 
of $A$ is equivalent to the vector problem penalizing sparsity of $\bx$, so we use 
the vector notation for simplicity. As both the Frobenius and nuclear norms are 
unitarily invariant\footnote{ Taking the SVD $A = U S V^T$
we have $\Vert U S V \Vert_F^2 = \Vert S \Vert_F^2$ and 
$\Vert U S V \Vert_* = \Vert S \Vert_*$ since $U$, $V$ are unitary.}, 
the analysis directly extends to the non-diagonal case.

The log heuristic for sparsity solves the following problem:
$\min \frac{1}{2}\Vert \bx - \by \Vert^2_2 + \alpha \sum_i \log (\delta + \vert x_i \vert)$, 
for a very small $\delta > 0$. This is a separable problem with a closed form 
solution for each coordinate\footnote{ For $\delta$ small enough, the global
minimum is always at $0$, but if $y > 2 \sqrt{\alpha}$ there is also a local minimum 
with a large domain of attraction between $0$ and $y$. 
Iterative linearization methods with small enough step size starting 
at $y$ will converge to this local minimum.
} (contrast this 
with the soft-thresholding operation):
{\footnotesize
\begin{equation}
	x_i = \begin{cases} \frac{1}{2}\left((y_i-\delta) + \sqrt{ (y_i-\delta)^2 - 4 (\alpha - y_i \delta)}\right), 
			~ y_i > 2 \sqrt{\alpha} \\
		\frac{1}{2} \left( (y_i+\delta) - \sqrt{ (y_i+\delta)^2 - 4 (\alpha+y_i \delta)} \right), 
		~ y_i < -2 \sqrt{\alpha}\\
		0, \mbox{ otherwise} \end{cases}
\end{equation}
}
Assuming that $\delta$ is negligible, then we have:
\begin{equation}
	x_i \approx \begin{cases} \frac{1}{2} ( y_i + \sqrt{ y_i^2 - 4 \alpha}), 
			\mbox{ if }~ y_i > 2 \sqrt{\alpha} \\
		 \frac{1}{2}(y_i - \sqrt{ y_i^2 - 4 \alpha}), 
		\mbox{ if }~ y_i < -2 \sqrt{\alpha}\\
		0, \mbox{ otherwise,} \end{cases}
\end{equation}

\begin{figure}[t]
      \centerline{\includegraphics[width=2.25in]{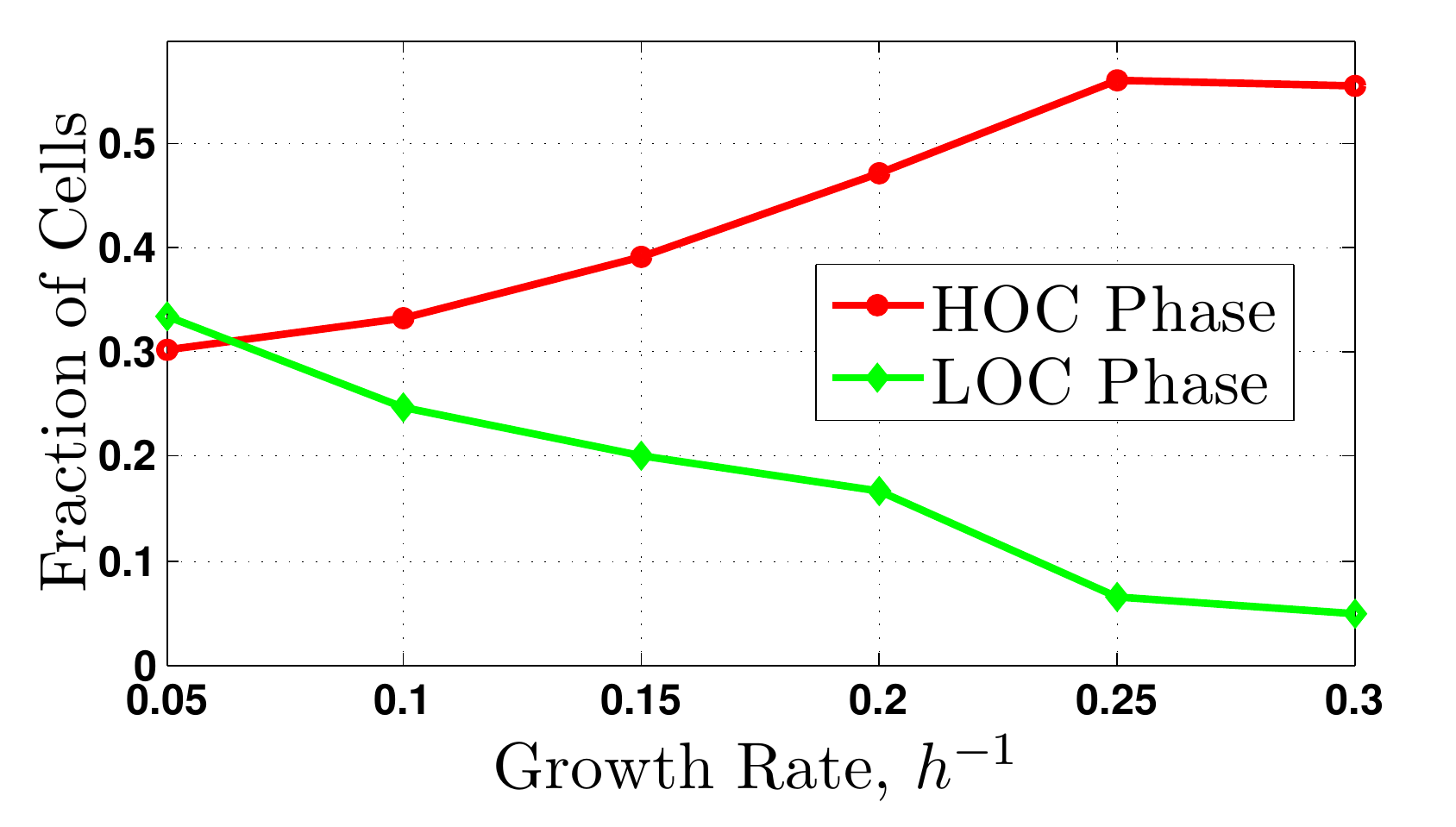}}
    \caption{\label{fig:bio_example} STLS infers accurate fractions of cells in 
	different physiological phases from measurements of population-average gene 
	expression across growth rate.}
	\vspace{-0.2in}
\end{figure}

and we chose $\alpha$ to annihilate the smallest entry in 
$\bx$, i.e. $\alpha = \frac{1}{4} \min_i y_i^2$. Sorting the entries in 
$|\by|$ in increasing order, with $y_0 = y_{\min}$, and defining 
$a_i = \frac{|y_i|} {|y_0|}$, we have $a_i \ge 1$ and the error in approximating 
the i-th entry, for $i > 0$ is 
\begin{equation}
Err_i = |x_i - y_i|^2 = \frac{y_0^2}{2} \left( a_i - \sqrt{a_i^2 - 1} \right)^2 \le \frac{y_0^2}{2 a_i^2}.
\end{equation}
Also, by our choice of $\alpha$, we have $Err_0 = y_0^2$ for $i = 0$. The approximation 
error quickly decreases for larger entries. 
In contrast, for $\ell_1$ soft-thresholding, the errors of approximating large entries are 
as bad as the ones for small entries. This analysis extends directly to the log-det 
heuristic for relaxing matrix rank.

\section{Conclusions}

We considered a convex relaxation for a very rich class of structured TLS problems, and 
provided theoretical guarantees. We also developed an efficient first-order augmented 
Lagrangian multipliers algorithm for reweighted nuclear norm STLS, which can be applied
beyond TLS to matrix completion and robust PCA problems. We applied STLS to quantifying cellular 
heterogeneity from population average measurements. In future  work we will study STLS 
with sparse and group sparse solutions, and explore connections to robust 
LS \cite{el_ghaoui_robust_LS97}.

\newpage
\bibliography{StructTLS}
\bibliographystyle{icml2014}

\end{document}